\documentclass{bmvc2k}

\title{360-Degree Gaze Estimation\\ in the Wild Using Multiple Zoom Scales}

\addauthor{Ashesh}{ashesh276@gmail.com}{1}
\addauthor{Chu-Song Chen }{chusong@csie.ntu.edu.tw}{1}
\addauthor{Hsuan-Tien Lin}{htlin@csie.ntu.edu.tw}{1}

\addinstitution{
 National Taiwan University\\
 Taipei, Taiwan
}

\runninghead{Ashesh, Chen, Lin}{Using Multiple Zoom Scales for Gaze Estimation}

\newcommand{\myparagraph}[1]{\vspace*{.25em} \noindent \textbf{#1}}

\begin{document}

\maketitle

\begin{abstract}
Gaze estimation involves predicting where the person is looking at within an image or video. 
Technically, the gaze information can be inferred from two different magnification levels: 
face orientation and eye orientation. The inference is not always feasible for gaze estimation 
in the wild, given the lack of clear eye patches in conditions like extreme left/right gazes or occlusions. In this work, we design a model that mimics humans' ability to estimate the gaze by aggregating from focused looks, each at a different magnification level of the face area. The model avoids the 
need to extract clear eye patches and at the same time addresses another important issue of face-scale variation for gaze estimation in the wild. We further extend the model to handle the challenging task of 
360-degree gaze estimation by encoding the backward gazes in the polar representation
along with a robust averaging scheme. Experiment results on the ETH-XGaze 
dataset, which does not contain scale-varying faces, demonstrate the model's effectiveness to assimilate information 
from multiple scales. For other benchmark datasets with many scale-varying faces (Gaze360 and RT-GENE), 
the proposed model achieves state-of-the-art performance for gaze estimation when using either images or videos. Our code and pretrained models can be accessed at \url{https://github.com/ashesh-0/MultiZoomGaze}.
\end{abstract}

\section{Introduction}
\label{sec:intro}
Gaze estimation is a critical task in computer vision for understanding human intentions and has significant potential to be used in unconstrained environments---i.e., in the
wild. Saliency detection~\cite{parks_augmented_2015,rudoy_learning_2013}, human-robot interaction~\cite{palinko_robot_2016, moon_meet_2014}, virtual reality~\cite{patneyanjul_towards_2016,padmanaban_optimizing_2017} are some of 
its main applications. The goal of gaze estimation is to predict the gaze direction of a given person accurately. The predicted direction can be
2D~\cite{krafka_eye_2016,huang_tabletgaze:_2016}, like the $(x, y)$ coordinate
of a mobile or laptop screen, or 3D~\cite{sugano_learning-by-synthesis_2014,wang_generalizing_2019}, reflecting the real-world reference system. 

\begin{figure}[t]
\centering
\includegraphics[width=0.95\textwidth]{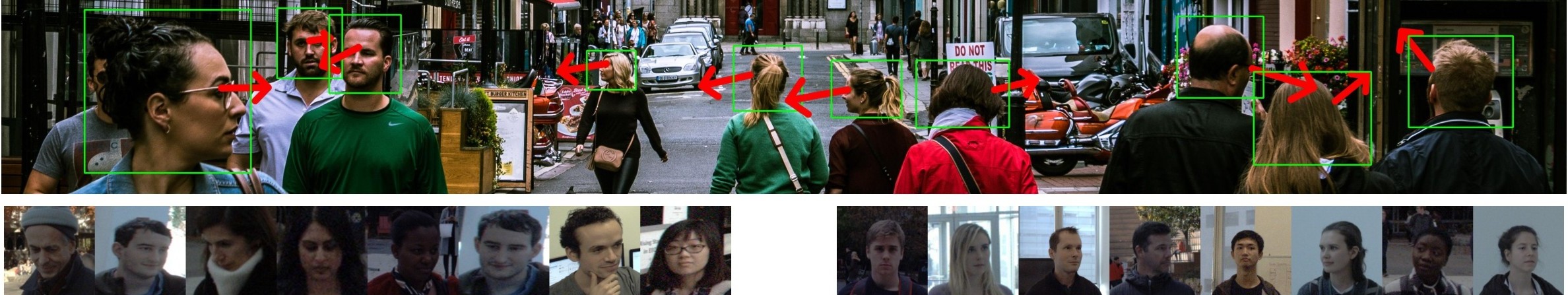}
\caption{(Top) Our model MSA produces qualitatively good gaze prediction (red arrow) in the wild with backward gazes and varying head sizes. Image taken from \url{www.pexels.com/photo/photo-of-people-walking-on-street-2416653/ }. (Bottom) Head crops from Gaze360 dataset. 
The head area in the image is referred to as the “scale” of the input image in this paper. 
Compared to right image set, left set has larger scale.}
\label{fig:DifferentScales}
\end{figure}

This work focuses on 3D gaze estimation. As shown in Figure~\ref{fig:DifferentScales}(Top), the task of 3D gaze estimation in the wild has two unique properties that make it challenging. Firstly, there can be occlusions, extremely left/right face orientations, and even backward gazes. For those subjects, details of the face and eye regions are often incomplete or totally absent. These two regions, each at a different magnification level, arguably carry most information for the gaze. Even when eye patches are available, they are of varying resolution because of the varying camera-person distance.
Recent approaches that leverage the face and eye information for gaze
estimation~\cite{chen_appearance-based_2019,cheng_coarse--fine_2020} generally rely on separate and therefore \textit{complete} and \textit{high-resolution} patches of the face and eyes. Therefore, those approaches are not suitable for our task of gaze estimation in the wild. Secondly, images in the wild setting have varying scales, where ``scale'' informally means the size of the head region. Varying scales come from varying camera-person distances and also result in varying resolutions of the subjects. Such variations cannot be easily
handled by usual convolutional neural networks 
(CNNs)~\cite{jaderberg_spatial_2015,lenc_understanding_2019}. 
In particular, we observe that the test performance varies significantly with varying scale. 

One simple approach to tackle the scale variation issue is to use a bounding box to crop and resize the head region, where the bounding box can be generated from a head detector, 
a face detector or a facial keypoint detector. 
For example, Kellnhofer et al.~\cite{kellnhofer_gaze360:_2019} work with the head crops of 
the raw images. However, producing a tight head-bounding box in the wild is challenging because of significant variations in the scale, background, head orientations and lighting conditions. 
Therefore, scale variation still persists even after re-scaling the head-crops, as shown in Figure~\ref{fig:DifferentScales}(Bottom). 
The aforementioned issue of incomplete facial information also makes it hard to leverage face or facial keypoint detectors for obtaining the bounding box.
That is, approaches that re-scale with bounding boxes do not fully solve the scale variation issue in this setting.

Most approaches have addressed the scale variation by leveraging image warping~\cite{sugano_learning-by-synthesis_2014} to normalize the input
image~\cite{fischer_rt-gene:_2018, zhang_its_2017, zhang_appearance-based_2015,
park_few-shot_2019,cheng_appearance-based_2018}. 
Those image-warping approaches re-orient the camera virtually such that the head orientation and the camera-person distance to the virtual camera can be fixed~\cite{zhang_its_2017,zhang_appearance-based_2015,park_few-shot_2019}.
But image warping typically requires locating the midpoint of the eyes accurately, and the midpoint is not always available under the aforementioned issue of incomplete facial information. That is, image-warping approaches are not satisfactory for gaze estimation in the wild, either.

In this paper, we tackle the two challenges of gaze estimation in the wild, namely extracting information from multiple magnification levels with incomplete facial information and significant scale variations, from a different perspective. 
Instead of bounding and normalizing the head region to \textit{one} scale in advance, we seek to mimic what humans do when trying to accurately estimate a gaze: take successively focused
looks at the head (face) region. In particular, we expand an input image to siblings scaled with different zooms, and aggregate the 2D feature maps from all siblings to make the final estimation. During feature extraction, every feature-map has an expression in all scales. Then, spatial max-pooling is applied on multiple 2D feature maps to get one 2D feature map of the same dimension, which reflects the intent to pick the best expression for each feature. The aggregation with multiple scales costs a mild increase of model complexity, but significantly improves the model by augmenting and aggregating with domain-justified transforms (zoom-in). Those transforms are simpler than the ones used in image-warping approaches, thus avoiding the complication of locating the midpoint between the eyes. In addition, aggregating multiple scales \textit{within} the model instead of resizing to one scale \textit{in advance} arguably makes the proposed model more robust than bounding-box-based approaches.

Another key challenge that we identify for estimation in the wild is 
the ambiguous representation of $-\pi$ and $\pi$ for the yaw angle that causes discontinuity in the parametric space.
For example, the Gaze360 dataset~\cite{kellnhofer_gaze360:_2019} contains backward gazes, for which the magnitude of the yaw angle is
greater than $\pi/2$. To resolve the ambiguity, we propose to encode the yaw angle using the complex exponential (polar) representation. This allows estimating the yaw angle from two different trigonometric functions. We analyze the two estimates and propose a weighted averaging scheme that takes the best of them to further improve the prediction.

The contributions of the paper can be summarized as follows: 1) To the best of our knowledge, we are the first to 
extract information from multiple scales for solving gaze estimation in the wild. Our strength is in the 
simplicity of our approach---not requiring sophisticated external modules for extracting eye patches, head pose or facial keypoints. Instead, our approach works with a simple head detector that does not have to be perfectly tight. The simplicity makes it possible to easily couple the approach across multiple backbones and different input types (images or videos).
2) To improve backward gaze prediction, which is critical for estimation in the wild, we propose to use
the polar representation
along with a robust averaging scheme
to estimate the yaw angle.
3) Our proposed approach achieves state-of-the-art performance on
Gaze360~\cite{kellnhofer_gaze360:_2019}, a benchmark dataset in the wild and RT-GENE~\cite{fischer_rt-gene:_2018}, another dataset in the wild with a natural environment.
We also demonstrate the effectiveness of the approach on ETH-Xgaze~\cite{zhang_eth-xgaze:_2020}, a dataset collected with a controlled environment.

\section{Proposed Model}

We first present our model for single-image input gaze estimation with full \(360^\circ\) variations in the yaw angle. We then extend the model to a more complicated task of video (sequential) gaze estimation and then to a simpler task where backward gazes are absent.

\myparagraph{Problem Formulation.} 
The state-of-the-art work for gaze estimation in the wild~\cite{kellnhofer_gaze360:_2019} converts the raw images (or video frames) in the wild, like Figure~\ref{fig:DifferentScales}(Top), to head-crop images using an off-the-shelf head detector to construct the training and test data. We follow the same construction. The construction makes each (head-crop) image roughly face-centered.

Given an input head-crop image~$I$, the 3D gaze estimation task for images aims
to predict the ground-truth yaw angle
\(\theta_{g}\) and pitch angle \(\phi_{g}\) of the gaze direction of the subject within $I$. We denote the predicted yaw and pitch as \(\theta_p\) and \(\phi_p\) respectively. 
The corresponding task for videos, identical to~\cite{kellnhofer_gaze360:_2019} takes a sequence of video frames \(V_{0:2T} = \{I_0,I_1,...,I_{T-1},
I_T,..,I_{2T}\}\) as the input and aims to predict the yaw and pitch angles for the frame \(I_T\). 

For evaluation, we convert the target gaze and its prediction from polar co-ordinate 
representation (\(\theta,\phi\)) to a unit vector in 3D cartesian co-ordinates. We 
then evaluate on angular error, which is the angle between the target and predicted unit 
vectors. 

\myparagraph{Target Domain Transformation.}
A gaze is said to be
a backward gaze when \(\theta_{g} \in \{[-\pi,-\pi/2]\) \( \cup [\pi/2,\pi] \} \). The model proposed in~\cite{kellnhofer_gaze360:_2019} directly
predicts \(\theta\) and \(\phi\), which we discovered to be causing considerable
losses on backward gazes due to a discontinuity in
the yaw domain: the yaw value jumps discontinuously from $\pi$ to $-\pi$. Refer supplemental for details. We tackle this problem via
target space transformation: we predict \(\sin(\theta)\), \(\cos(\theta)\), and
\(\sin(\phi)\). We use tanh activation to ensure apppropriate prediction range.

To estimate \(\theta\) from the predicted \(\sin(\theta)\) and \(\cos(\theta)\),
we use a two-step method. First, we estimate \(\theta\) in two ways. In the
sine-based way, we estimate yaw \(\theta_S\) from \(\sin(\theta)\) and
\(\mathrm{sign}(\cos(\theta)\));%
\footnote{
$\mbox{sign}(\cdot)$ returns $+1$ on a non-negative number, and $-1$ otherwise.
}
in the cosine-based way, we estimate yaw
\(\theta_C\) using \(\cos(\theta)\) and \(\mathrm{sign}(\sin(\theta)\)). 

In the second step, we calculate our final estimate of yaw, \(\theta_p\). Initially we used \(\theta_p =
\theta_{SC}\) where \(\theta_{SC}=(\theta_{S} + \theta_{C})/2 \). However, 
\(\theta_S\) is much more accurate than \(\theta_C\) around
\(0^{\circ}\) which can be observed in the distribution of
\(\theta_{g}\), \(\theta_S\), and \(\theta_C\) shown in 
supplemental. From the dip in the distribution of \(\theta_C\)
around \(0^{\circ}\) primarily and that of \(\theta_S\) slightly around \(\pm 90^\circ\), one can say that model 
is having difficulty in predicting those regions. We argue that
the low derivative of the \(\tanh\) activation function near \(\pm 1\) makes it
difficult for the model to predict values very close to \(\pm 1\), 
and that the high derivative of the \(\sin^{-1}\) and \(\cos^{-1}\) functions around 1
discourages the prediction of angles close to \(\pm 90^\circ\) and \(0^\circ\)
respectively. We address this using a weighted averaging scheme.
Specifically, our yaw prediction \(\theta_{\mathit{WSC}}\) is defined as
\(\theta_{\mathit{WSC}}=w*\theta_{S} + (1-w)*\theta_{C}    \). Here, \(w\) is defined as \(w = |\cos((\theta_{S} + \theta_{C})/2) | \).
Defined this way, in practice, \(\theta_S\) is assigned greater weight when
\(\theta_{g}\) is near $0^\circ$ and \(\theta_C\) is assigned greater weight when
\(\theta_{g}\) is near \(\pm \pi/2\).

\myparagraph{Loss Function.}
Following the Gaze360 paper~\cite{kellnhofer_gaze360:_2019}, we use 
Pinball loss for all experiments on the dataset. In generic terms, if \(y_{g}\) is the target
and \(y_p\) is prediction, then the quantile loss \(L_\tau\) for quantile
\(\tau\) is defined as 
\begin{align}
  L_\tau(y_p, \sigma,y_{g}) = \max({\tau}{y_\tau}, -(1-\tau)y_\tau) &&
    y_{\tau} = \begin{cases}
    y_{g} - (y_p -\sigma) & \text{for } \tau \le 0.5\\
    y_{g} - (y_p +\sigma) & \text{otherwise}
\end{cases}  
\end{align}

Here, \(\sigma\) can be understood as the uncertainty in prediction, which is yet
another output from our network. This formulation is used on all three gaze
targets, namely \(\sin(\theta)\), \(\cos(\theta)\), and \(\sin(\phi)\). The final loss
is the average of these losses over quantiles \(\tau=0.1\) and \(\tau=0.9\).
See \cite{kellnhofer_gaze360:_2019} for more details. 

\begin{figure}[t]
\centering
\includegraphics[width=0.85\textwidth]{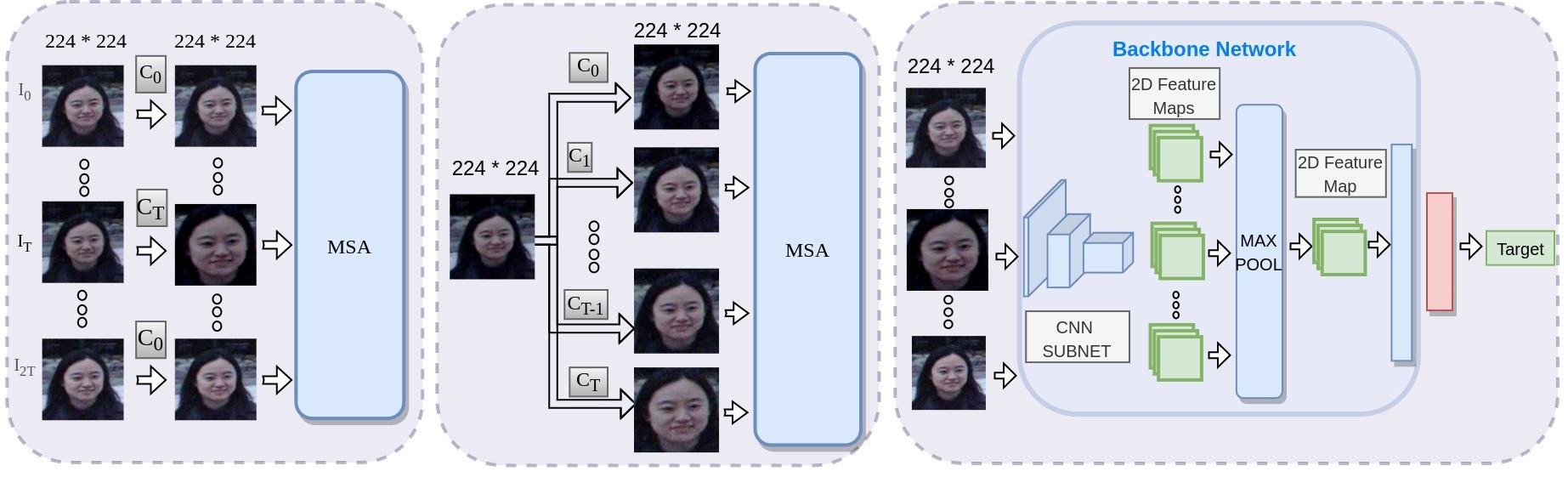}
\caption{ MSA (right) and data pre-processing for static (middle) and sequence (left) model. }
\label{fig:architecture}
\end{figure}

\myparagraph{Architecture for Static Model.} In Gaze360~\cite{kellnhofer_gaze360:_2019}, as shown in
Figure~\ref{fig:DifferentScales} (Bottom), head crops have varying
scales. Additionally, the pinball static 
model~\cite{kellnhofer_gaze360:_2019} together with the proposed sine-cosine transformation does not 
handle images of varying scales any better (refer to the
Experiments section for details). We use center-cropping along with
spatial max-pooling to handle scale robustly and extract multi-scale features.

The architecture of the proposed MSA (Multiple Scale Aggregation) model and data preprocessing 
for single image input case is shown in Figure~\ref{fig:architecture} (Middle and Right). 
The input image is center-cropped with multiple sizes
\(\mbox{CCropL} = \{C_i, i \in [0,T]\}\) and subsequently scaled back to the original
size, yielding a set of images with varying scales, which are then
fed into the proposed MSA module. It comprises of a pre-trained backbone network and 
a final dense layer. Images are fed into the CNN portion of the backbone network to 
produce 2D feature maps for differently scaled
images.  We then use max-pooling on these stacked 2D feature maps along the
scale dimension, and pass the output thereof through the backbone head, which is
comprised of dense layers. Finally, the output from the backbone head is passed
through a dense layer and an appropriate activation to yield the predictions. To
predict \(\sin(\theta)\), \(\cos(\theta)\),  and \(\sin(\phi)\), we use tanh
activation. We then use the \(\theta_{\mathit{WSC}} \) formulation to get \(\theta\). 
To predict \(\sigma\), we use sigmoid activation.

\myparagraph{Architecture of Sequence Model.}
With sequence model, we use MSA with a different data preprocessing as shown 
in Figure~\ref{fig:architecture} (Left). Given input
image sequence $I_0,I_1..I_{2T}$, we center-crop the images with sizes
$C_0,C_1..C_{T-1},C_T,C_{T-1}..C_1,C_0$ respectively and rescale them back to original size.
We also introduce the constraint that $C_i > C_{i+1} \forall i$. The specific crop-size ordering and
constraint is placed to implicitly preserve information about frame ordering. Notably, $C_T$, 
the crop size for the target frame $I_T$, is the smallest which gives the greatest zoom-in effect. 
This should encourage the model to extract more micro-level details from eye region in $I_T$. 
These rescaled images are then fed to MSA to get the prediction.

\myparagraph{Changes for Non-Backward Gaze Estimation.}
When solving the simpler non-backward gaze estimation, the absence of
yaw discontinuity means that the proposed sine-cosine transformation is not needed. 
Therefore, in this setting, we directly predict \(\theta\) and \(\phi\) which
essentially reduces the dimension of last dense layer from 4 to 3. 
We test this setting on the RT-GENE dataset~\cite{fischer_rt-gene:_2018} 
and ETH-Xgaze~\cite{zhang_eth-xgaze:_2020} dataset.

\section{Experiments}
\subsection{Implementation Details and Data}\label{subsec:data}
\myparagraph{Implementation Details} Besides using pretrained Resnet18~\cite{he_deep_2015} for the backbone as done in ~\cite{kellnhofer_gaze360:_2019}, MobileNet~\cite{sandler_mobilenetv2:_2019},
SqueezeNet~\cite{iandola_squeezenet:_2016},
ShuffleNet~\cite{ma_shufflenet_2018}, and Hardnet~\cite{chao_hardnet:_2019}  were also used. All weights get updated during training. Like~\cite{kellnhofer_gaze360:_2019}, $T=3$ was used for sequence model. For all experiments on all datasets and both model types, unless specified otherwise, we used 
[224,200,175,150] for CCropL, obtained empirically. More details given in supplemental.

\myparagraph{Data} We conducted experiments on Gaze360~\cite{kellnhofer_gaze360:_2019}, RT-GENE~\cite{fischer_rt-gene:_2018} and ETH-XGaze~\cite{zhang_eth-xgaze:_2020}. Gaze360 includes images of both indoor and outdoor scenes,
with a full $360^{\circ}$ range of yaw angles.
The camera-to-person distance varies considerably (1\,m to 3\,m) leading to variable head
sizes and image resolutions. Most existing datasets~\cite{smith_gaze_2013,
sugano_learning-by-synthesis_2014, mora_eyediap:_2014, krafka_eye_2016,
huang_tabletgaze:_2016, zhang_its_2017} don't contain this much variation in
camera-person distance and yaw angle. Using 238 subjects, this dataset comprises 129K
training images, 17K validation images, and 26K test images. Similar to~\cite{kellnhofer_gaze360:_2019}, we only work with head crops provided in the dataset.

RT-GENE~\cite{fischer_rt-gene:_2018} also has varying camera-person distances (80--280\,cm). 
RT-GENE includes four sets of images:
two original sets and two inpainted sets. There are two versions of both sets:
the raw version and a MTCNN~\cite{zhang_joint_2016} based normalized version.
As in \cite{cheng_gaze_2020}, we find the inpainted sets to be
noisy and therefore we only use the Raw-Original and Normalized-Original set. However, since the
normalized version used facial keypoints for rescaling the image, we note that those image sets are therefore not relevant for the 
unconstrained setting.

ETH-XGaze data covers a wide range of head poses, gaze angles and lighting conditions and has high resolution images. It has 1M images comprising of 80 subjects in train set and 15 in test set. While the pre-processed images, which we use, have fixed camera-person distance, the wide range of head poses and gaze angles makes it a useful dataset for evaluation of in the wild performance.

\begin{table}
\parbox{.5\textwidth}{
    \begin{tabular}{|p{0.23\textwidth}|p{0.04\textwidth}|p{0.04\textwidth}|p{0.04\textwidth}|}
    \hline
     Model + Backbone & All 360 & Front 180 & Back\\
    \hline\hline
    {Pinball Static \cite{kellnhofer_gaze360:_2019}} & 15.6 & 13.4 & 23.5 \\
    \hline
    {Pinball Static \cite{kellnhofer_gaze360:_2019} Re-Implemented} & 15.6 & 12.8 & 26.0 \\
    \hline
    Spatial Weights CNN~\cite{zhang_its_2017} & 20.7 & 16.8 & 34.9 \\
    \hline
    Spatial Weights CNN~\cite{zhang_its_2017} + Resnet & 15.5 & 12.8 & 25.4 \\
    \hline
    CA-Net~\cite{cheng_coarse--fine_2020} & 18.2 & 15.3 & 28.6 \\
    \hline
    CA-Net~\cite{cheng_coarse--fine_2020} + Resnet & 15.2 & 12.8 & 23.6 \\
    \hline
    Static+avg & 14.4 & 12.8 &  20.4\\
    \hline
    Static+wavg & 14.4 & 12.7 & 20.4 \\
    \hline
    {MSA+raw} & 15.8 & 12.4 & 28.1\\
    \hline
    {MSA+avg} & 14.0 & 12.3 & 19.9\\
    \hline
    {MSA} & \textbf{13.9} & \textbf{12.2} & \textbf{19.9}\\
    \hline
    \end{tabular}
}
\parbox{.5\textwidth}{
    \begin{tabular}{|p{0.22\textwidth}|p{0.04\textwidth}|p{0.04\textwidth}|p{0.04\textwidth}|}
    \hline
     Model & All 360 & Front 180 & Back\\
    \hline\hline
    Pinball LSTM \cite{kellnhofer_gaze360:_2019} & 13.5 & 11.4 & 21.1 \\
    \hline
    SSA+avg+Seq & 13.1 & 11.5 & 18.9 \\
    \hline
    SSA+avg+Seq + LSTM & 13.2 & 11.5 & 19.4 \\
    \hline
    SSA+wavg+Seq + LSTM & 13.1 & 11.4 & 19.3 \\
    \hline
    SSA+wavg+Seq & 13.0 & 11.4 & \textbf{18.8} \\
    \hline
    {MSA+avg+Seq + LSTM} & 12.7 & 10.9 & 19.2 \\
    \hline
    {MSA+avg+Seq} & \textbf{12.5} & 10.8 & 19.0 \\
    \hline
    {MSA+Seq + LSTM} & 12.7 & 10.9 & 19.2 \\
    \hline
    {MSA+Seq} & \textbf{12.5} & \textbf{10.7} & 19.0 \\
    
    \hline
    \end{tabular}
}
\caption{Performance comparison on Gaze360 dataset~\cite{kellnhofer_gaze360:_2019}. For each configuration, three models were trained. Average angular error is reported in this table. Resnet18 is used as backbone in our model and its variants. (Left) Comparison on Static models.  (Right) Comparison on Sequential (Temporal) models.}
\label{tab:gaze360}
\end{table}

\begin{figure}
\begin{tabular}{cccc}
\bmvaHangBox{\includegraphics[width=3cm]{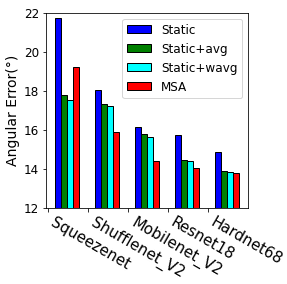}}&
\bmvaHangBox{\includegraphics[width=2.2cm]{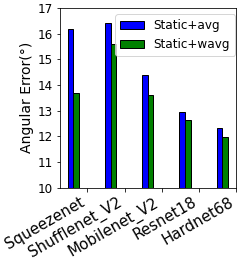}}&
\bmvaHangBox{\includegraphics[width=2.4cm]{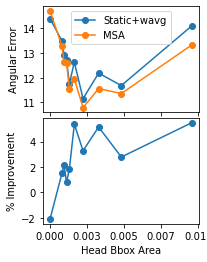}}&
\bmvaHangBox{\includegraphics[width=3.8cm]{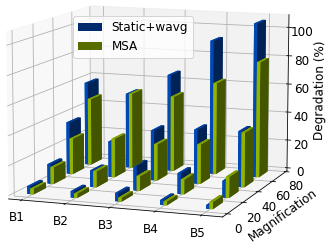}}\\
(a)&(b)&(c)&(d)
\end{tabular}
\caption{All results are computed on Gaze360 dataset. For (a), (b) and (d), B1, B2, B3, B4, and B5 denote Squeezenet, Shufflenet\_V2, Mobilenet\_V2, Resnet18, and
Hardnet68 backbones respectively.(a) Performance of Static model variants. (b) Angular error on samples having front $\pm 20^\circ$ as yaw. (c) Top: Angular error variation with variation in Head bounding bbox area (in fractions) on front $180^\circ$ images. Head bbox area is binned into 10 equal sized bins and average angular error is reported for each bin. Bottom: Percentage benefit of our MSA model over Static+wavg model with Head bbox area variation.(d) Percentage increase in average angular error on
front $180^\circ$ gazes with varying amount of magnification. }
\label{fig:perf_backbones}
\end{figure}

\begin{table}[ht]
\parbox{.67\textwidth}{
 \centering
 \begin{tabular}{|p{0.2\textwidth}|p{0.03\textwidth}|p{0.14\textwidth}|p{0.12\textwidth}|p{0.03\textwidth}|}
\hline
 Method & Bkb & Image & Input & Err\\
\hline
\hline
Spatial weights CNN~\cite{zhang_its_2017} & - & NOrig & Face & $10$ \\
\hline
RT-Gaze \cite{fischer_rt-gene:_2018} & - & NOrig & Eye &  $8.6$ \\
\hline
RT-Gaze \cite{fischer_rt-gene:_2018} & - & NOrig + NIn & Eye &  $7.7$ \\
\hline
FAR-Net \cite{cheng_gaze_2020} & - & NOrig & Face + Eye & $8.4$ \\
\hline\hline
Static & B4 & Orig & Face &  $7.9$\\
\hline
{MSA+raw} & B4 & Orig & Face & $7.1$\\
\hline
Static & B5 & Orig & Face & $7.0$ \\
\hline
{MSA+raw} & B5 & Orig & Face & \textbf{6.7} \\
\hline
Static & B4 & NOrig & Face & $7.2$\\
\hline
{MSA+raw} & B4  & NOrig  & Face & $7.3$\\
\hline
{MSA+raw}+175 & B4 & NOrig & Face & $6.9$\\
\hline
\end{tabular}
\caption{Performance comparison on RT-GENE dataset. NOrig, Orig and NIn stand for Normalized-Original, Raw-Original and Normalized-Inpainted image type respectively. B4 and B5 stands for Resnet18 and Hardnet68 respectively.}
\label{table:PerfRTGENE}
}
\hfill
\parbox{.3\textwidth}{
\centering
\begin{tabular}{|p{0.13\textwidth}|p{0.03\textwidth}|p{0.03\textwidth}|}
\hline
 Method & Bkb  & Err\\
\hline
ETH-XGaze~\cite{zhang_eth-xgaze:_2020} & -- & 4.5\\
Pinball Static~\cite{kellnhofer_gaze360:_2019} & B4 & 4.4\\
MSA+raw & B4 & 4.1\\
MSA+raw & B5 & 4.0\\
\hline
\end{tabular}
\caption{Performance comparison on ETH-XGaze dataset. B4 and B5 stands for Resnet18 and Hardnet68 respectively}
\label{table:PerfXGaze}
}

\end{table}

\subsection{Model Performance Comparison}
\myparagraph{Benchmarks.} Due to the challenging nature of the dataset, we could not find any other 
work on Gaze360 dataset apart from the original work~\cite{kellnhofer_gaze360:_2019} of which we directly 
report the performance from their paper. On implementing their Static model~\cite{kellnhofer_gaze360:_2019}, we found different performance in few yaw ranges and so we report our implementation as 'Pinball Static~\cite{kellnhofer_gaze360:_2019} Re-Implemented'. We adapt work of 
Cheng et al.~\cite{cheng_coarse--fine_2020} which also makes use of features from multiple scales 
by extracting features from eye region and face. We use the eye bounding boxes provided with the dataset for this work. We also adapt work of Zhang et al~\cite{zhang_its_2017} where, 
similar to us, they work with just face images. For both of above mentioned works, we also report results by 
replacing their face feature extractor (backbone network) with Resnet18 and the loss with pinball 
loss so as to have a more authentic comparison.

\myparagraph{Ablation Study.} We compare our MSA model with multiple variants of itself. Here, we 
mention differing points for each variant against our model.
MSA+avg and MSA uses the \(\theta_{SC}\) and \(\theta_{WSC}\) formulation for yaw respectively. MSA+raw directly 
predicts \(\theta\), \(\phi\) and uncertainty \(\sigma\), and does not use the sine-cosine 
transformation. Static+avg and Static+wavg do not use the multi-crop based idea and can be understood as being
MSA+avg and MSA respectively with \(\mbox{CCropL}=[224], T=1\).  SSA, used in sequential model is MSA with no
multicrop, i.e, \(\mbox{CCropL}=[224,224,224,224], T=3\).

\myparagraph{Performance.} In Table~\ref{tab:gaze360} (left) we compare Static model performance. Firstly, by comparing Static+avg model's performance ($20.4^\circ$) on Back gazes with ~\cite{kellnhofer_gaze360:_2019} ($23.5^\circ$), benefit of sine-cosine transformation can be seen.
Next, benefit of multi-scale aggregation can be seen on Front 180 gazes by observing MSA+raw's performance ($12.4^\circ$) with ~\cite{kellnhofer_gaze360:_2019}($13.4^\circ$).  Finally, MSA model outperforms on all three gaze categories namely front $180^\circ$--- $|\theta_{g}| \in [0^\circ,90^\circ]$ ($12.2^\circ$), back --- $|\theta_{g}| \in [90^\circ,180^\circ]$ ($19.9^\circ$) and overall ($13.9^\circ$). It is worth noting the inferior performance of ~\cite{cheng_coarse--fine_2020} (overall $15.2^\circ$) which used both eye and face patches. It shows that absence of clear eye patches in such conditions significantly hampers the performance.

In Table~\ref{tab:gaze360} (right), we show that in Sequence model type, our model MSA+Seq outperforms the benchmarks on all three gaze categories. We also show
the performance on using the LSTM module as an aggregation module instead of our max-pool operator for a closer comparison to \cite{kellnhofer_gaze360:_2019}. More details are given in the supplementary material, where we do an experiment of varying the aggregation modules.

\myparagraph{Individual Effect of Different Components.}
Figure~\ref{fig:perf_backbones} (a) shows the performance gains
achieved by different components. We observe the benefit of the sine-cosine
transformation by comparing Static with Static+avg, and observe the benefit of
multiple zoom scales by comparing MSA with Static+wavg. 
We argue that reason for inferior performance of MSA with Squeezenet was that it being the lightest backbone (1.2M parameters), could not manage multiple scales. Finally, as \(\theta_{\mathit{WSC}}\) was introduced to fix issues with \(\theta_{SC}\) around primarily $0^\circ$, its benefit can be observed
on frontal gazes in Figure~\ref{fig:perf_backbones} (b).

\myparagraph{Performance Variation With Head Crop Area.}
In Gaze360 dataset, the authors provide the head bounding box dimensions as a fraction of the original fixed size full body image. Firstly, as seen in Figure~\ref{fig:perf_backbones} (c), for front $180^\circ$ gazes, we find considerable variation in performance with change in head bounding box area. Secondly, MSA gets better performance consistently on most bounding box area bins. 

\myparagraph{Performance Variation with Camera-Person Distance}
We look at how much MSA outperforms the Static+wavg model on front 180 gazes as we vary the distance. We binned images on camera-person distance into three equal bins. We show percentage improvement achieved by MSA over Static+wavg in Table~\ref{table:improvment_with_distance}. MSA gives more benefit for larger distances. 

\myparagraph{Performance Degradation with Scale Perturbations.}
Here we quantify performance degradation if a zoomed-in/zoomed-out image is given instead as input at evaluation. For every
magnification level, we created one zoomed-in image and one zoomed-out image. For
a magnification of $c$, we center-cropped the original image with size $224
-c$ and rescaled it back to size 224 to create a zoomed-in image. To produce a 
zoomed-out image, we padded the original image with $c/2$ pixels on the boundary 
and rescaled it down to 224. Thus, the greater the magnification $c$, the greater the 
zoom-in and zoom-out effect. In Figure~\ref{fig:perf_backbones} (d), we plot the
percentage increase in angular error (averaged over zoom-in and zoom-out) with the
amount of magnification on front $180^\circ$ gazes. In the majority of cases across backbones
and magnification levels, MSA has lower percentage
increment in angular error as compared to Static+wavg thereby showing robustness of multi-zoom approach on scale variations.

\myparagraph{On Time Complexity and GPU RAM}
MSA+Seq takes same time and RAM as Pinball LSTM~\cite{kellnhofer_gaze360:_2019}. MSA takes L times more time w.r.t Pinball Static~\cite{kellnhofer_gaze360:_2019}, L being the number of crop sizes. However, unlike LSTM, MSA can be parallelized 
since the 2D feature map computation for each magnification level is independent and so MSA+Seq can run faster than Pinball LSTM~\cite{kellnhofer_gaze360:_2019} and MSA can achieve comparable speed to Pinball Static~\cite{kellnhofer_gaze360:_2019}. 
If we parallelize MSA, then it however would take L times more GPU RAM over Pinball Static~\cite{kellnhofer_gaze360:_2019}. Otherwise, 2 times GPU RAM is needed for which one would need to compute feature map for each cropsize one by one while simultaneously updating the max feature map stored in a buffer. But since all our experiments needed just 4 crop-sizes, we argue that even for Static model type, this is not a big limitation for MSA. MSA with Resnet18 evaluates 279 images in 1s wall time and takes 2.5GB GPU RAM on a single 2080 Ti GPU with 64 batch size and 4 workers. With 1 batch size and 1 worker, it is 80 images/sec. 

\myparagraph{Comparison with feature-map upscaling}
One way to get a faster model could be to rescale feature map instead of rescaling input image. Inspired from~\cite{Hausler_2021_CVPR}, here, we passed the image through the initial portion of the backbone network (Resnet18) to get a 2D feature map. We then center-cropped the feature map with different sizes and rescaled them back. We obtain these cropsizes by transforming CCropL. For 14*14 feature map, $C_i$ crop-size will transform to $\frac{C_i*14}{224}$. These rescaled feature-maps are then averaged and passed through the remaining network to get the prediction. We used  \(\theta_{\mathit{WSC}} \) formulation. From Table~\ref{table:fmap_upscaling}, we find feature-map upscaling to be worse than MSA for all feature-map resolutions.

\begin{figure}[t]
\centering
\includegraphics[width=0.85\textwidth]{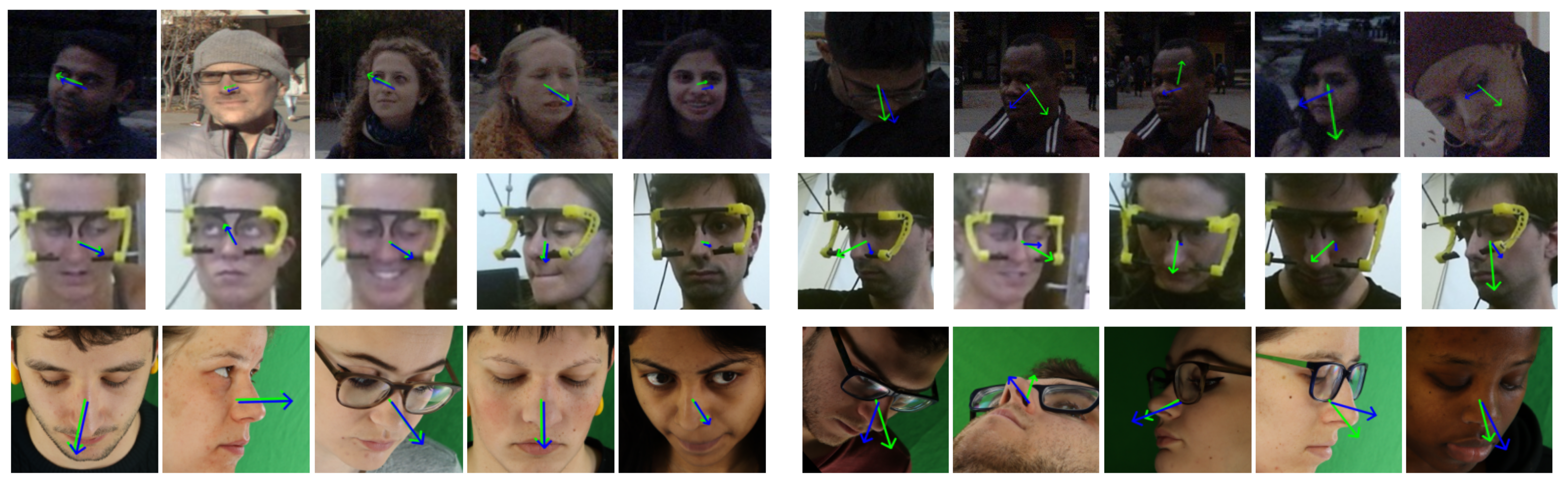}
\caption{ Showing some of the better (Left) and worse (Right) performing images for MSA model on Gaze360 (Row 1), RT-GENE (Row 2) and ETH-Xgaze (Row 3) dataset. Green and purple arrow denote the ground truth and predicted gaze directions respectively. }
\label{fig:best_worst}
\end{figure}

\myparagraph{Qualitative Analysis} In Figure~\ref{fig:best_worst}, we look at few of the better and worse performing images from all three datasets. One could see that in cases where (1) the head orientation is highly oblique and when (2) eyeball is deviated from center position but is less visible, the model performs poorly. Performance on some backward gazes were also worse but we didn't include them here since there is little visual cue to glean from them. Model naturally performs best on relatively more frequent head poses.
\begin{table}[]
\parbox{.5\textwidth}{
 \centering
 \begin{tabular}{|p{0.1\textwidth}|p{0.1\textwidth}|p{0.11\textwidth}|p{0.1\textwidth}|}
\hline
 Res & All 360 & Front 180\\
\hline
$7*7$   & 14.5 & 12.8\\
$14*14$ & 14.5 &12.8 \\
$28*28$ & 14.5 &12.6 \\
\hline
\hline
\end{tabular}
\caption{Performance of upscaling based idea on Gaze360 dataset. First column, Res stands for feature map resolution}
\label{table:fmap_upscaling}
}
\hfill
\parbox{.45\textwidth}{
\centering
\begin{tabular}{|p{0.14\textwidth}|p{0.19\textwidth}|}
\hline
 Distance(m) & \% Improvement\\
\hline
0.9-1.8 & 3.9\%\\
1.8-2.7 & 4.5\%\\
2.7-3.5 & 5.2\%\\
\hline
\end{tabular}
\caption{Variation of \% improvement of MSA over Static+wavg model w.r.t camera-person distance on Gaze360.}
\label{table:improvment_with_distance}
}

\end{table}

\myparagraph{Model performance on RT-GENE and ETH-XGaze}
In Table~\ref{table:PerfRTGENE}, we see that MSA+raw gets state-of-the-art results on
RT-GENE. We directly report the performance of Spatial
weights CNN and RT-Gaze on RT-GENE from \cite{fischer_rt-gene:_2018}.
Likewise for FAR-Net from \cite{cheng_gaze_2020}.  However, MSA+raw($7.3^\circ$) with default CCropL does not outperform Static model($7.2^\circ$) with Normalized-Original (NOrig) images. This 
makes sense because unlike Raw-Original, NOrig images do not have
significant variations in scale and also are a bit zoomed in.
Consequently, the proposed minimum cropsize of 150 in CCropL proves too low 
and results in inferior performance. However, when CCropL is set to \([224,208,191,175]\) (MSA+raw+175 model), which has the higher min-cropsize 175, it outperforms Static model. As mentioned in Section~\ref{subsec:data}, comparing on Normalized-Original is unfair to the in the wild setting on which we focus on. Next, Table~\ref{table:PerfXGaze} compares the performance on ETH-XGaze dataset. 
It is important to note that here, we worked with normalized images--- the camera-head distance is fixed. In absence of 
scale variation, the outperformance of MSA+raw model over Static pinball model~\cite{kellnhofer_gaze360:_2019} points to 
its ability to extract information from multiple scales.

\section{Related Work}
Initial work on gaze estimation was model-based~\cite{guestrin_general_2006,jianfeng_eye-model-based_2014,
wang_hybrid_2016,wang_deep_2016}, involving modeling the geometry of
the eye and then using this for gaze estimation. With the rise of
computational resources, the increase in dataset sizes, and the emergence of deep
learning, appearance-based
approaches~\cite{zhang_appearance-based_2015,cheng_appearance-based_2018,wang_hierarchical_2018,park_few-shot_2019,fischer_rt-gene:_2018} came to being. 
As the name suggests, this approach involves estimating the gaze
directly from the image appearance. A typical appearance-based model is a
neural network which takes as input an image containing a human face or an
eye patch and then predicts the gaze from this. Unlike the model-based approach, here the
setup is quite practical and the environment more relaxed. Most 
appearance-based models require a single camera with lower requirements for 
high-resolution images. However, since this does not capture eye geometry, it is
highly susceptible to changes in head pose. The fact that several
appearance-based approaches take the head pose as input along with the
image~\cite{funes_mora_gaze_2012,lu_head_2012,lu_head_2011,sugano_learning-by-synthesis_2014}
is a testament to the sensitivity of this approach to head pose changes.

Initial appearance-based models worked with eye patches~\cite{zhang_appearance-based_2015}. Subsequently, face images began to be
used as input~\cite{zhang_its_2017}. More recently, both face and eye patch have been used together 
as input, demonstrating the presence of useful information at multiple scales~\cite{chen_appearance-based_2019,cheng_coarse--fine_2020}.                      %
As discussed before, one problem with this is the need for bounding boxes for
the eyes, especially when dealing with large gaze angles, low-resolution images, or occlusions. 

As with the development in models, there has been a gradual evolution of datasets. With recent
datasets we see a larger number of
participants~\cite{krafka_eye_2016}, greater variation in head pose and gaze 
angle~\cite{fischer_rt-gene:_2018,kellnhofer_gaze360:_2019}, and greater
variation in camera-person distance, background variations, and so on.
Notably, the Gaze360 dataset~\cite{kellnhofer_gaze360:_2019} has
several interesting properties. With full $360^\circ$ variation in yaw,
significant camera-person distance variations, and both indoor and
outdoor background settings, Gaze360 is a promising dataset for 3D gaze
estimation in the wild.
In the context of scale, we also find the RT-GENE
dataset~\cite{fischer_rt-gene:_2018} to be quite useful, due to its
significant variation in camera-person distance (80--280\,cm).

\section{Conclusion}
We present a novel approach for gaze estimation in the wild where we extract information from different magnification levels by aggregating features from images belonging to these levels. Our work is simple and can be easily coupled with different input types (images or videos). Furthermore, the approach reaches state-of-the-art performance for gaze estimation in the wild without relying on more complicated components like eye patch detector, face pose detector, or image warping. The multi-zoom approach has potentials for other vision problems in the wild as well, where the scale variation is a crucial issue.
\section*{Acknowledgement}
We thank the anonymous reviewers for valuable suggestions. This work is partially supported by the Ministry of Science and Technology of Taiwan via the grants MOST 107-2628-E-002-008-MY3, 110-2628-E-002-013, and 110-2218-E-002-033-MBK. We also thank the National Center for High-performance Computing (NCHC) of National Applied Research Laboratories (NARLabs) in Taiwan for providing computational resources.
\bibliography{egbib}
\end{document}


\maketitle
\begin{figure}[h]
    \centering
    \includegraphics[width=\textwidth]{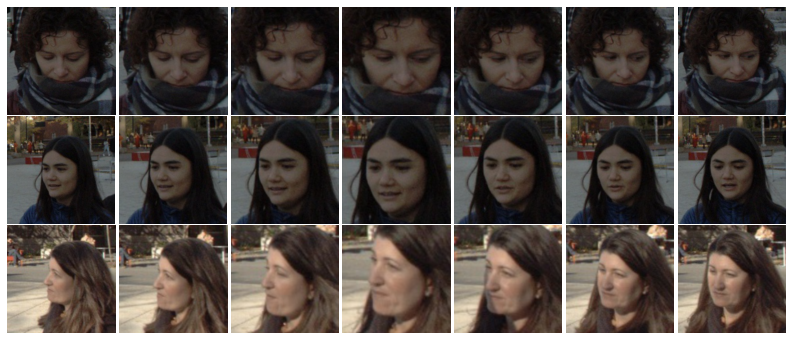}
    \caption{We show 3 pre-processed image sequences for our Sequential (Video) model, one in each row. With T=3 set for our experiments, sequence length becomes $2T+1$ = 7. The target frame, the frame for which we want to predict the gaze is the $4^{th}$ frame. Firstly, one can notice that change in eye movement is quite significant across frames. Therefore, one would ideally want eye level details to come majorly from the target frame. This is achieved by our pre-processing technique. Note, how images are successively zoomed-in in first 4 images and then zoomed out in last 4 images in this figure. This  gives maximum zoom-in effect to the $4^{th}$ frame. Secondly, this specific zoom-in zoom-out ordering also implicitly encodes the information about ordering of the frames in the input sequence--- the most zoomed-in frame is the target frame 
    and the more zoomed-out an image seems with respect to the target frame, the farther apart it is from it in time. Here, Frame 4 is most zoomed-in and with respect to it, Frame 1 is more zoomed out than Frame 2 is which in turn is more zoomed out than Frame 3.
    This is important specifically because we want to predict gaze 
    for the $4^{th}$ frame but our aggregation technique, the spatial max-pooling does not care about ordering of the sequence. }
    \label{fig:seq_imgs}
\end{figure}

\begin{figure}[h]
    \centering
    \includegraphics[width=\textwidth]{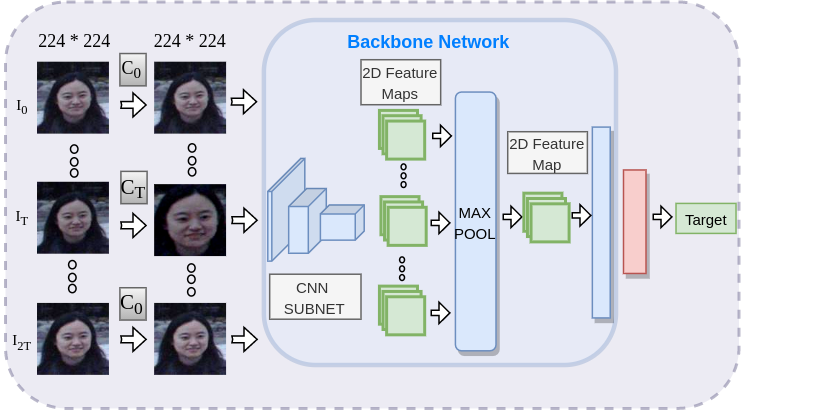}
    \caption{An end-to-end schema for Sequential (Video) model. $2T+1$ frames are center-cropped and rescaled 
    to original size according to the scheme mentioned in the main text. This scheme ensure greatest zoom-in 
    affect for the middle frame. Since head pose changes are relatively minor when compared to eye movements 
    over these frames, one would want eye orientation to get majorly captured from the middle frame, the frame 
    for which we want to predict the gaze. Middle frame getting greatest zoom-in ensures larger and detailed 
    availability of eye region thereby satisfying our objective.}
    \label{fig:seq_detailed}
\end{figure}
\section{Overview}
Here, we briefly enumerate the content present in different sections. In Section~\ref{sec:seq}, we give
experiment backed intuition on why we chose the specific pre-processing scheme for our sequential (video) model MSA+Seq.
In Section~\ref{sec:yawdiscontinuity}, we provide more details regarding the discontinuity in Yaw angle for full $360^\circ$ 
yaw prediction. We also present data showing the limitation of $\theta_{SC}$ formulation due to which we came up with $\theta_{WSC}$.
In Section~\ref{sec:limitations}, we comment on few limitations of our work. We give implementation details in Section~\ref{sec:implementation}. Next, we present a couple of more experiments 
including ones on varying aggregation module and using LSTM in Section~\ref{sec:experiments}. Finally, we provide the standard error for 
data presented in Table 1 in the main manuscript and values and standard errors for the plots in the main manuscript in the later sections.

\section{Pre-processing Scheme for Sequential Model }\label{sec:seq}
Our sequential (Video) model  uses the MSA architecture with a specific 
pre-processing scheme. Let us call the frame for which we want to predict the gaze as target frame. Let us also call our proposed pre-processing scheme as ZoomIn. 
The scheme ensures that maximum zoom-in effect is applied on
the target frame. It also ensures that the sequence order
information of the frames gets implicitly encoded--- prediction is to be done for most zoomed-in
frame and the less zoomed-in a frame is, farther it is from the target frame in terms of sequence order. 
One can see few examples pre-processed with ZoomIn scheme in Figure~\ref{fig:seq_imgs}.

The benefit of this scheme can be seen in a slightly different problem definition for video gaze prediction. 
In this formulation, instead of the middle frame, last frame is chosen as the target frame. 
In this case, according to our scheme, maximum zoom-in must be applied to the last frame. Similar to
the default configuration, we have same T ($T=3$) and CCropL. But now, given input
image sequence $I_0,I_1,I_2,I_3,I_4,I_5,I_{6}$, we center-crop the images with sizes
$C_0,C_0,C_1,C_1,C_2,C_2,C_3$ respectively and rescale them back to original size. Here, 
$C_i$ is the $(i+1)^{th}$ element in CCropL. 
We compare this with two other zoom-in schemes. Firstly, we use a 'Random' zoom-in
scheme. In this, we permute the centercrop sizes [$C_0,C_0,C_1,C_1,C_2,C_2,C_3$] randomly before applying them for every sequence, both in
training and in evaluation. This way, the MSA model has no way to get  the frame-ordering information. Secondly,
we use a 'Reverse' zoom-in scheme where the target frame  gets the least zoomed-in effect.
In this case, given input image sequence $I_0,I_1,I_2,I_3,I_4,I_5,I_{6}$, we center-crop the images with sizes
$C_3,C_2,C_2,C_1,C_1,C_0,C_0$ respectively and rescale them back to original size.
As can be seen in Table~\ref{table:video_future_prediction}, our configuration ZoomIn performs best. 
Significantly better performance of our ZoomIn scheme and Reverse scheme over Random scheme shows that 
implicitly present sequence information helps ZoomIn and Reverse schemes. 
Better performance of ZoomIn over Reverse scheme shows the advantage of 
highest zoom-in effect given to the target frame.
\begin{table}
\parbox{.5\textwidth}{
\centering
\begin{tabular}{|p{0.3\textwidth}|p{0.15\textwidth}|}
\hline
 Pre-processing Scheme & Angular error\\
\hline
None & 13.0 \\
Random & 12.6 \\
ZoomIn & 12.6 \\
Reverse & 12.6 \\
\hline
\end{tabular}
\caption{Performance of MSA+Seq model with varying Pre-processing Scheme. First row corresponds to SSA+wavg+Seq model. We predict gaze for the \textit{middle} frame of the sequence}
\label{table:suppl_ordering}
}
\hfill
\parbox{.45\textwidth}{
\centering
\begin{tabular}{|p{0.2\textwidth}|p{0.1\textwidth}|}
\hline
 Pre-processing Scheme & Angular error\\
\hline
Random & 16.0 \\
Reverse & 13.4 \\
ZoomIn & 13.2 \\
\hline
\end{tabular}
\caption{Performance of MSA+Seq model with varying Pre-processing Scheme. Here, 
we predict the gaze for the \textit{last} frame.}
\label{table:video_future_prediction}
}
\end{table}

We also did the same experiment on our original problem definition where the target frame is the 
middle frame. In this case, we find that our scheme, the reverse scheme and the random scheme are all giving 
very similar performance as can be seen in Table~\ref{table:suppl_ordering}. This has to do with how 
Gaze360 dataset was created. It contains mostly monotonic movement of target gaze across frames. People
start from a gaze orientation and move their gaze consistently in one direction. Due to this, the average
gaze of the seven frames will naturally be very close to the gaze of the middle frame, which is the target gaze. So, the 
network just needs to obtain average gaze of all the frames and for this, the sequence information does not matter. 
Relatively higher importance to the eye region of the target frame also does not matter when one aims to 
take the average of all gazes. 

\section{Discontinuity In Yaw}\label{sec:yawdiscontinuity}
\begin{figure}
\begin{tabular}{cc}
\bmvaHangBox{\includegraphics[width=0.5\textwidth]{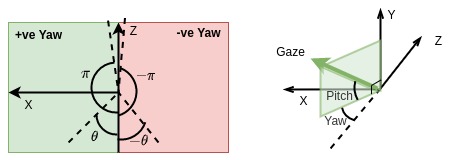}}&
\bmvaHangBox{\includegraphics[width=0.4\textwidth]{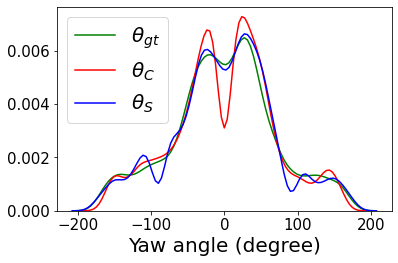}}\\
(a)&(b)
\end{tabular}
\caption{(a) Discontinuity in yaw space. Yaw angle defined with respect to
negative z axis (right figure). Discontinuity (left figure) is seen when the projection 
of the gaze vector on the XZ
plane is very close to the positive Z direction. From one side, the angle reaches
$\pi$, and from the other, it reaches $-\pi$.
(b) Distribution of actual and predicted yaw angle. Note the dip around 0$^\circ$ 
in the distribution of the cosine-based estimate \(\theta_C\)}
\label{fig:DiscontInYaw}
\end{figure}

 We want to elaborate on two points on this subject. Firstly, as stated in the main manuscript, 
 backward gazes not only are harder to estimate owing to absence
of face in the input image, but also introduce discontinuity in the yaw angle. 
Specifically, for small $\epsilon$, backward gazes of yaw angles $\pi-\epsilon$ and $-\pi+\epsilon$ are far apart in terms of the numerical value but are of close physical proximity as can be seen in Figure~\ref{fig:DiscontInYaw} (a). 

Secondly, in Figure~\ref{fig:DiscontInYaw} (b), we show the distribution of $\theta_{C},\theta_{S}$ and the ground truth $\theta_{gt}$. One can observe a significant dip at around $0^\circ$ in $\theta_C$. This indicates that $\theta_C$ has difficulty in predicting around $0^\circ$ and which arguably is the cause for inferior performance of $\theta_{SC}$ on frontal $\pm 20^\circ$. The reason for this and the remedy of using $\theta_{WSC}$ is given in the main manuscript.

\section{Challenges and Limitations}\label{sec:limitations}
We observe that the MSA model does not give benefit to backward gazes. This becomes clear
 when one looks at the Back column of Table 1 of the main manuscript. We feel this is intuitive.
 MSA aims to extract information from two magnification scales--- eye region and overall head region.
 For backward gazes, the face is not visible--- eyes, nose and other facial details are completely absent.
 So, one needs to estimate gaze from the overall head orientation, i.e, the head pose. As the information
 is not contained in multiple magnification levels, the MSA approach gives no benefit. 
 
 The other limitation of this approach is time complexity for single-image input model type. 
Firstly, we note that for sequential models, MSA+Seq takes same amount of time as SSA+avg+Seq and so 
 there is no issue of time complexity here. For single-image input models, the time complexity of MSA 
 scales linearly with the number of elements in CCropL. CCropL=[$224,200,175,150$] takes 4 times more 
 time than Static model. As far as the GPU RAM is concerned, MSA takes twice GPU RAM as Pinball 
 Static~\cite{kellnhofer_gaze360:_2019}. For working within twice GPU RAM, one would need to compute 
 the 2D feature map one by one while simultaneously updating the max feature map stored in a buffer.
 CCropL=[$224,150$] takes two times more time than Static model. However, unlike LSTM, 
 MSA can be parallelized since the 2D feature map computation for each magnification level is 
 independent of each other and so one could easily improve upon the time complexity issue. 
 
\section{Implementation Details}\label{sec:implementation}
Similar to~\cite{kellnhofer_gaze360:_2019}, we fixed the backbone's output layer size to
256 for all our experiments. Our network input is 224x224x3-sized images.
For having a fair comparison with~\cite{kellnhofer_gaze360:_2019}, we also used $T=3$ for sequence model. 
For all experiments on all three datasets and both model types 
(video and single-image), unless specified otherwise, we used 
[224,200,175,150] for CCropL, which we obtained empirically. When using the 
LSTM module as an aggregation module,
following \cite{kellnhofer_gaze360:_2019}, we used bidirectional LSTM with two
layers and a hidden size of 256.  We implemented the model using PyTorch. For all
of the Gaze360 experiments, we trained the network for 100 epochs with a batch size of 64,
a learning rate of 0.0001, and the Adam optimizer. 

For the experiments on the RT-GENE dataset~\cite{fischer_rt-gene:_2018},
following their GitHub code,\footnote{\url{https://github.com/Tobias-Fischer/rt_gene/}}
we used a learning rate of 0.000325 and the Adam optimizer with the same hyperparameters
($\mathit{betas}=(0.9, 0.95)$). Additionally, we use early stopping on
validation loss with a patience of 5.

\section{Additional Experiments}\label{sec:experiments}
\subsection{Empirical Evidence on Significance of Scale in Gaze360 dataset} \label{subsec:gaze360_scale_variation}
Here, we use Static+avg model with the MobileNet backbone.
At evaluation time, we added center-cropping with a fixed size and subsequent resizing to the original size
in the data preprocessing step for both the test and validation sets. Note that during 
model training, no center-cropping was performed on the training and validation sets.
One therefore expects performance to degrade upon the introduction
of center-cropping at evaluation time. However, we found a significant proportion of images having 
a better performance with the cropping based preprocessing so much so that, we observe a minute performance improvement
on overall dataset as can be seen in 
Table~\ref{table:EmpiricalEvidOnScale}. This implies that dataset contains images of varying scales and the model does not
extract features equally efficiently from all scales. We observe similar findings on RT-GENE dataset as 
well whose data is presented in next subsection.

\begin{table}
\parbox{.45\linewidth}{
\begin{tabular}{|c|c|c|}
\hline
 Centercrop size & Val error & Test error\\
\hline\hline
224 & 13.72 & 13.83 \\
210 & 13.68 & 13.78 \\
200 & 13.73 & 13.76 \\
\hline
\end{tabular}
\caption{Angular error obtained on front $180^\circ$ gazes in Test and
Validation sets using Static+avg model with MobileNet backbone on Gaze360
dataset. At evaluation time, center-crop followed by resize operation is done.}
\label{table:EmpiricalEvidOnScale}
}
\hfill
\parbox{.45\linewidth}{
\begin{tabular}{|c|c|}
\hline
Aggregation module & Angular error\\
\hline
Spatial-Max & 13.9 \\
MAX & 14.1 \\
Spatial-Attention & 14.0 \\
LSTM & 14.1\\
\hline
\end{tabular}
\caption{Comparison of different aggregation modules with MSA}
\label{table:AggregationModVariation}
}
\end{table}

\subsection{Empirical Evidence on Significance of Scale in RT-GENE dataset}
As shown in Table~\ref{table:ScaleRTGENE}, similar to Gaze360,
overall performance on the validation set improves when we add
center-crop preprocessing at evaluation time for RT-GENE dataset. Being more prominent for K-fold=0,
this holds true for all three folds with the Hardnet68 backbone. This indicates that
(1) images in the data have multiple scales and that (2) the Static model~\cite{kellnhofer_gaze360:_2019} does not
capture features from all scales equally well. It is worth noting that this is an 'indicator' experiment. For us to take cue from it, it is not necessary for this effect to manifest with all backbones to the extent that the average performance improves. 

\begin{table}
\centering
\begin{tabular}{|c|c|c|c|c|c|c|}
\hline
\multirow{}{}{Center-crop size} & \multicolumn{3}{|c|}{Val error (Hardnet68)} & \multicolumn{3}{|c|}{Val error (Resnet18)} \\
\cline{2-7}
 & K-fold 0 & K-fold 1 & K-fold 2 & K-fold 0 & K-fold 1 & K-fold 2\\
 \hline
 224 & 7.21 & 5.68 & 5.86 & 7.67 & 6.14 & \textbf{6.38} \\
 215 & 7.04 & \textbf{5.66} & \textbf{5.84} & \textbf{7.60} & \textbf{6.14} & 6.4 \\
 210 & \textbf{7.00} & 5.68 & 5.88 & 7.62 & 6.20 & 6.49 \\
 \hline
\end{tabular}
\caption{Angular error on Validation set
using static model with Resnet18 and Hardnet backbones on RT-GENE}\label{table:6}
\label{table:ScaleRTGENE}
\end{table}

\subsection{Benefit of Using Multiple-Scales Over Single Optimal Scale}
Given the evidence in Subsection~\ref{subsec:gaze360_scale_variation} showing that Gaze360 dataset has decent variations in scale, it is reasonable to assume the existence of a single centercrop size which would give better performance.  We therefore also wanted to check whether our MSA+avg model performs better than Static+avg model with that optimal center crop size. To be specific, all input images will be center cropped to that crop size and they will then be subsequently rescaled to original size of 224. Rescaled images are then fed to Static+avg model. We did the experiment with Resnet18 backbone. Results present in Table \ref{table:SupplOptimalCropSize} show that 175 as centercrop size is the optimal configuration. As can be seen from Table 1 of the main manuscript, our MSA+avg model outperforms it. Note that the comparison done this way is not fair for our MSA+avg model as one cannot know optimal centercrop size for test dataset a priori. Inspite of this, this outperformance shows that using multiple cropsizes is better over the use of one crop size.

\begin{table}
\parbox{.4\textwidth}{
\begin{tabular}{|c|c|}
\hline
 Centercrop size & Angular error\\
\hline
224 & \(14.5 \pm 0.16\) \\
200 & \(14.31 \pm 0.06\) \\
175 & \(14.3 \pm 0.13\) \\
150 & \(14.4 \pm 0.19\) \\
\hline
\end{tabular}
\caption{Finding the optimal crop-size for Static+avg + Reg model with Resnet18 backbone.}
\label{table:SupplOptimalCropSize}
}
\strut
\parbox{.5\textwidth}{
\begin{tabular}{|p{0.25\textwidth}|p{0.08\textwidth}|p{0.1\textwidth}|}
\hline
 Method & Backbone & Angular Error\\
\hline
{MSA+raw + LSTM} & Resnet & $7.1$\\
{MSA+raw + LSTM} & Hardnet & $6.9$ \\
\hline
\end{tabular}
\caption{ Performance using LSTM as aggregation module on RT-GENE dataset using Raw-Original image type}
\label{table:suppl_rtgene}
}
\end{table}

\begin{table*}
\centering
\begin{tabular}[width=\columnwidth]{|c|c|c|c|c|c|}
\hline
Name & Backbone & All 360 &	Front 180&	Front 40 &	Back \\
\hline
Spatial Weights CNN~\cite{zhang_its_2017} & - & 0.28 & 0.27 & 0.33 & 0.34 \\
Spatial Weights CNN~\cite{zhang_its_2017} & Resnet & 0.18 & 0.13 & 0.21 & 0.56 \\
CA-Net~\cite{cheng_coarse--fine_2020} & - & 0.54 & 0.49 & 0.34 & 1.77 \\
CA-Net~\cite{cheng_coarse--fine_2020} & Resnet & 0.18 & 0.20 & 0.19 & 0.10 \\
Static+avg & Resnet & 0.06 &	0.04 &	0.18 & 0.12 \\
Static+wavg & Resnet & 0.06 & 0.05 & 0.12 & 0.13 \\
MSA+raw & Resnet & 0.10 & 0.13 & 0.38 & 0.48 \\
MSA+avg & Resnet & 0.18 & 0.27 & 0.58 & 0.16 \\
MSA & Resnet & 0.17 & 0.25 & 0.50 & 0.12 \\
MSA & Hardnet &0.08 & 0.13 & 0.31 & 0.089\\
\hline
\end{tabular}
\caption{Standard Error of models presented in Table 1 (Left) of main manuscript}
\label{table:SupplStaticSem}
\end{table*}

\begin{table*}
\centering
\begin{tabular}[width=\columnwidth]{|c|c|c|c|c|c|}
\hline
Name & Backbone & All 360 &	Front 180&	Front 40 &	Back \\
\hline
SSA+avg & Resnet & 0.05 & 0.04 & 0.06 & 0.36 \\
SSA+wavg+Seq & Resnet & 0.02 & 0.04 & 0.06 & 0.24 \\
MSA+avg+Seq + Reg + LSTM & Resnet & 0.06 & 0.08 & 0.22 & 0.31\\
MSA+Seq + Reg + LSTM & Resnet & 0.05 & 0.07 & 0.17 & 0.29\\
MSA+avg+Seq & Resnet & 0.06 & 0.05 & 0.12 & 0.19\\
MSA+Seq & Resnet & 0.08 & 0.06 & 0.13 & 0.24\\
MSA+Seq & Hardnet & 0.05 & 0.08 & 0.17 & 0.06\\
\hline
\end{tabular}
\caption{Standard Error of models presented in Table 1 (Right) of main manuscript}
\label{table:SupplSeqSem}
\end{table*}

\begin{figure}
\centering
\includegraphics[width=\textwidth]{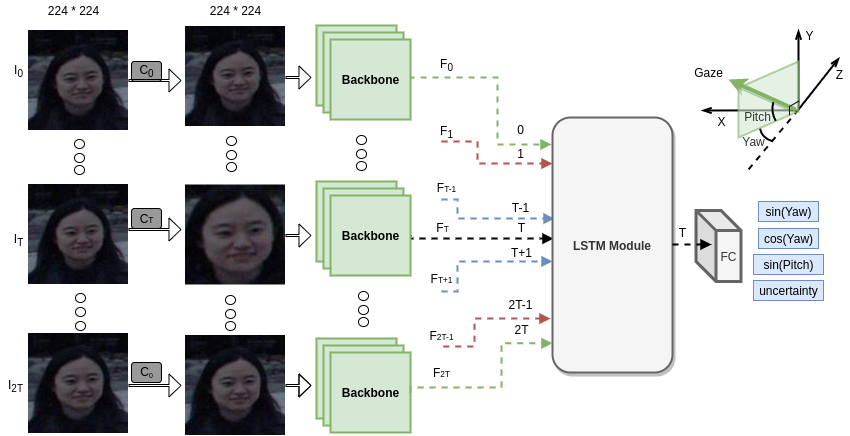}
\caption{A variant of our Seq[W]SCMulticrop model which has LSTM as aggregation module}
\label{fig:LSTMSeqModel}
\end{figure}

\subsection{Effect of Different Aggregation Modules}
Here, we varied the aggregation modules: we used Spatial-Max (used in
our proposed model), LSTM, MAX and Spatial-Attention as aggregation techniques.
For Spatial-Attention, 2D feature maps were aggregated using an attention
module instead of max-pool. For LSTM and MAX, one-dimensional features coming out of the backbone network's last
dense layer were aggregated. For LSTM, the input sequence ordering followed
CCropL. For MAX, maximum was taken along the scale dimension. We used Resnet18 as the backbone for this study. As seen in
Table~\ref{table:AggregationModVariation}, performance is not significantly dependant on the 
choice of the aggregation technique.

\subsection{LSTM as an Aggregation Module}
We did extensive experiments with LSTM as an aggregation module. As briefly described in the main document, the output of the backbone network is passed into the LSTM. This is different from the aggregation module for our main model MSA, since in that case, output of the CNN portion of the backbone network is aggregated. For more clarity, please refer to Figure~\ref{fig:LSTMSeqModel}. We did the experiments on both Sequence and Static model. For the Static model, output of LSTM module corresponding to last element of input sequence is taken and passed through dense layers to yield the target variables. For the Sequence model, output of the LSTM module corresponding to the middle element of the input sequence is taken and passed through the dense layers to yield the target. It is done this way for the sequence model since we want to predict the gaze for the middle frame in the input sequence.
\begin{table}
\parbox{.45\textwidth}{
    \begin{tabular}{|c|p{0.1\textwidth}|}
    \hline
    Aggregation Module & Standard Error \\
    \hline
    SPATIAL-MAX & 0.18 \\
    MAX & 0.10 \\
    SPATIAL-ATTENTION & 0.08 \\
    LSTM & 0.11 \\
    \hline
    \end{tabular}
    \caption{Standard Error of models presented in Table~\ref{table:AggregationModVariation}}
    \label{table:SupplMultiAggregation}
}
\end{table}

\paragraph{Regularized Pinball Loss:}
As stated in the manuscript, following Gaze360 paper~\cite{kellnhofer_gaze360:_2019}, we keep our loss as Pinball Loss. We explored another way here to fix model's bad performance near \(\theta=0^\circ\). We added a constraint penalizing deviation of predicted \(\sin(\theta)\), \(\cos(\theta)\) from  \(\sin^2(\theta) + \cos^2(\theta) = 1\). It is implemented as addition of a weighted MSE loss component. Our final loss is $$w*L_{P} + (1-w)*L_{MSE}$$, where $L_P$ is the pinball loss,  $L_{MSE}=MSE(1,sin^2(\theta) + cos^2(\theta))$ and $w$ set to $0.9$. We denote the presence of this regularization by ``Reg'' token in model name in the tables. We found this to slightly outperform with a couple of backbones (majorly with Squeezenet). One can verify the same by looking at performance data for this configuration in Tables \ref{table:lstm_sc}, \ref{table:lstm_wsc}, \ref{table:lstm_multicropsc} and \ref{table:lstm_multicropwsc}. However, one can see that even with the regularization, the weighted sine-cosine transformation \(\theta_{WSC}\) still gives better performance over naive sine-cosine transformation \(\theta_{SC}\) . We, therefore did not include it in our final model configuration. Nonetheless, one can observe the benefit of using our Static[W]SCMultiCrop model over Static[W]SC model from above tables.

Performance on Static and Sequence model types are shown in Table \ref{table:suppl_static} and Table \ref{table:suppl_seq} respectively for Gaze360 dataset. Results on RT-GENE dataset are shown in Table \ref{table:suppl_rtgene}. 

\begin{center}
\begin{table*}
\centering
\begin{tabular}{|p{0.3\textwidth}|p{0.13\textwidth}|p{0.13\textwidth}|p{0.13\textwidth}|p{0.13\textwidth}|}
\hline
 Model &  All 360 & Front 180 & Front Facing & Back\\
\hline
{MSA+avg + LSTM + Reg}  & \(14.1\pm 0.10\) & \(12.3\pm 0.10\) & \(12.4\pm 0.20\) &\(20.6\pm0.38\)\\
{MSA + LSTM + Reg }  & \(14.1\pm0.11\) & \(12.3\pm 0.10\) & \(12.3\pm0.20\) & \(20.6\pm0.37\)\\
\hline
\end{tabular}
\caption{ Performance comparison for Static models with LSTM as aggregation module and with Resnet backbone on Gaze360 dataset \cite{kellnhofer_gaze360:_2019} }
\label{table:suppl_static}
\end{table*}
\end{center}

\begin{center}
\begin{table}
\centering
\begin{tabular}{|p{0.32\textwidth}|p{0.14\textwidth}|p{0.13\textwidth}|p{0.13\textwidth}|p{0.13\textwidth}|}
\hline
 Model &  All 360 & Front 180 & Front Facing & Back\\
\hline
\hline
SSA+avg+Seq + LSTM + Reg  & \(13.18\pm0.03\) & \(11.45\pm0.06\) & \(11.2\pm0.11\) & \(19.4\pm0.34\)\\
SSA+wavg+Seq + LSTM + Reg  & \(13.12\pm0.05\) & \(11.39\pm0.05\) & \(10.9\pm0.19\) & \(19.3\pm0.35\)\\
{MSA+avg+Seq + LSTM + Reg} & \(12.74\pm 0.06\) & \(10.94\pm0.08\) & \(10.7\pm0.22\) & \(19.2\pm0.32\) \\
{MSA+Seq + LSTM + Reg} & \(12.71\pm0.05\) & \(10.91\pm0.07\) & \(10.6\pm0.17\) & \(19.2\pm0.30\) \\
\hline
\end{tabular}
\caption{ Performance comparison for Sequential models with LSTM as aggregation module and with Resnet backbone on Gaze360 dataset \cite{kellnhofer_gaze360:_2019} }
\label{table:suppl_seq}
\end{table}
\end{center}

\section{Standard Error Data on Model Performances}
In Table \ref{table:SupplStaticSem}, \ref{table:SupplSeqSem} and \ref{table:SupplMultiAggregation} , we present the data containing standard error of models present in Table 1 (Left) and Table 1 (Right) of of the main manuscript and Table~\ref{table:AggregationModVariation} respectively. We train the models 3 times and compute the standard deviation of mean angular error.
\begin{table}
    \begin{tabular}{|p{0.12\textwidth}|c|c|c|c|c|}
    \hline
    Backbone & Static & Static+avg & Static+wavg & MSA+avg & MSA \\
    \hline
    Squeezenet & \(20.9\pm 0.89\) & \(17.8 \pm 0.10\) & \(17.49\pm0.03\) & \(20.2\pm0.18\) & \(19.62\pm 0.37\) \\
    Shufflenet & \(17.96\pm 0.06\) & \(17.25\pm 0.04\) & \(17.12\pm 0.07\) & \(16.1\pm0.11\) & \(15.9 \pm 0.12\) \\
    Mobilenet & \(16.3\pm 0.19\) & \(15.8\pm 0.47\) & \(15.7\pm0.41\) & \(14.6\pm0.13\) & \(14.5\pm0.10\) \\
    Resnet18 & \(15.78\pm0.07\) & \(14.44\pm0.06\) & \(14.35\pm0.06\) &\(14.0\pm0.18\) & \(13.9\pm0.17\)\\
    Hardnet68 & \(14.7\pm0.18\) & \(13.91\pm0.05\) & \(13.85\pm0.06\) & \(13.8 \pm0.10\) & \(13.72\pm0.08\) \\
    \hline
    \end{tabular}
    \caption{Model Performances (with standard error) on all $360^\circ$ gazes}
    \label{table:raw_all360}
\end{table}

\begin{table}
    \begin{tabular}{|p{0.12\textwidth}|c|c|c|c|c|}
    \hline
    Backbone & Static & Static+avg & Static+wavg & MSA+avg & MSA \\
    \hline
    Squeezenet & \(15.2\pm0.46\) & \(15.4\pm0.17\) & \(15.12\pm0.08\) &\(17.8\pm0.28\) &\(17.3\pm0.30\)\\
    {Shufflenet} & \(15.2\pm0.10\) & \(15.55\pm0.04\) & \(15.42\pm0.07\) &\(14.13\pm0.06\) & \(14.02\pm0.07\)\\
    {Mobilenet} & \(13.8\pm0.19\) & \(14.1\pm0.53\) & \(14.0\pm0.49\) & \(12.9\pm0.10\) & \(12.8\pm0.12\)\\
    Resnet18 & \(13.0\pm0.13\) & \(12.77\pm0.04\) & \(12.69\pm0.05\) & \(12.3\pm0.27\) & \(12.20\pm0.25\)\\
    Hardnet68 & \(12.4\pm0.10\) & \(12.3\pm0.10\) & \(12.2\pm0.11\) & \(12.10\pm0.14\) & \(12.03\pm0.13\)\\
    \hline
    \end{tabular}
    \caption{Model Performances (with standard error) on Front $180^\circ$ gazes}
    \label{table:raw_front180}
\end{table}
\begin{table*}
    \centering
    \begin{tabular}{|c|c|c|c|c|c|}
    \hline
    Backbone & Static & Static+avg & Static+wavg & MSA+avg & MSA \\
    \hline
    Squeezenet & \(13.3\pm0.56\) & \(16.5\pm0.96\) & \(13.6\pm0.1\) &\(20.6\pm0.62\) &\(13.5\pm0.25\)\\
    Shufflenet & \(15.1\pm0.1\) & \(16.45\pm0.09\) & \(15.63\pm0.06\) &\(14.2\pm0.18\) &\(13.2\pm0.16\)\\
    Mobilenet  & \(13.79\pm0.05\) & \(14.48\pm0.44\) & \(13.68\pm0.12\) &\(12.9\pm0.22\) &\(12.27\pm0.19\)\\
    Resnet18     & \(13.1\pm0.25\) & \(13.1\pm0.18\) & \(12.8\pm0.12\) & \(12.6\pm0.58\) &\(12.1\pm0.50\) \\
    Hardnet68    & \(12.3\pm0.37\) & \(12.4\pm0.11\) & \(12.08\pm0.11\) & \(12.1\pm0.40\)&\(11.67\pm0.30\)\\
    \hline
    \end{tabular}
    \caption{Model Performances (with standard error) on front facing ($40^\circ$) gazes}
    \label{table:raw_front40}
\end{table*}
\begin{table*}
    \centering
    \begin{tabular}{|c|c|c|c|c|c|}
    \hline
    Backbone & Static & Static+avg & Static+wavg & MSA+avg & MSA \\
    \hline
    Squeezenet & \(41\pm2.7\) & \(26.3\pm0.24\) & \(26.0\pm0.34\) & \(28.9\pm0.61\) & \(28.2\pm0.54\)\\
    Shufflenet & \(28.0\pm0.43\) & \(23.1\pm0.5\) & \(23.2\pm0.13\) & \(23.0\pm0.50\) & \(22.9\pm0.58\) \\
    Mobilenet & \(25.5\pm0.35\) & \(22.0\pm0.76\) & \(21.7\pm0.7\) & \(20.9\pm0.29\) & \(20.6\pm0.25\)\\
    Resnet18 & \(25.6\pm0.21\) & \(20.4\pm 0.12\) & \(20.4\pm0.13\) & \(19.9\pm0.16\) & \(19.9\pm0.12\) \\
    Hardnet68 & \(22.9\pm0.50\) &\(19.7\pm0.21\) & \(19.6\pm0.17\) & \(19.98\pm0.04\) & \(19.81\pm0.09\)\\
    \hline
    \end{tabular}
    \caption{Model Performances (with standard error) on back gazes}
    \label{table:raw_back}
\end{table*}
\begin{table*}
    \centering
    \begin{tabular}{|c|c|c|c|c|}
    \hline
    Model &  All 360 & Front 180 & Front Facing & Back\\
    \hline
    Squeezenet & 15.94 &	13.05 &	12.85 &	26.32 \\
    Shufflenet & 16.11 &	14.31 &	14.94 &	22.61 \\
    Mobilenet & 14.71 &	12.86 &	12.87 &	21.34 \\
    Resnet18 & 14.07 &	12.19 &	12.14 &	20.86 \\
    Hardnet68 & 13.65 &	11.98 &	11.93 &	19.69 \\
    \hline
    \end{tabular}
    \caption{Angular error with LSTM as aggregation module for MSA+avg+LSTM+REG model}
    \label{table:lstm_multicropsc}
\end{table*}

\begin{table*}
    \centering
    \begin{tabular}{|c|c|c|c|c|}
    \hline
    Model &  All 360 & Front 180 & Front Facing & Back\\
    \hline
    Squeezenet & 15.91 &	13.03 &	12.64 &	26.27 \\
    Shufflenet & 16.07 &	14.26 &	14.67 &	22.58 \\
    Mobilenet & 14.66 &	12.83 &	12.71 &	21.24 \\
    Resnet18 & 14.05 &	12.16 &	12.04 &	20.84 \\
    Hardnet68 & 13.63 &	11.95 &	11.85 &	19.66 \\
    \hline
    \end{tabular}
    \caption{Angular error with LSTM as aggregation module for MSA+LSTM+REG model}
    \label{table:lstm_multicropwsc}
\end{table*}

\begin{table*}
    \centering
    \begin{tabular}{|c|c|c|c|c|}
    \hline
    Model &  All 360 & Front 180 & Front Facing & Back\\
    \hline
    Squeezenet & 16.72 &14.26 &	14.74 &	25.57 \\
    Shufflenet & 17.52 & 15.91 &	16.49 &	23.3 \\
    Mobilenet & 15.48 &	13.74 &	13.14 &	21.76 \\
    Resnet18 & 14.41 & 12.66 & 12.81 & 20.72 \\
    Hardnet68 & 13.81 & 12.19 & 12.29 & 19.64 \\
    \hline
    \end{tabular}
    \caption{Angular error with LSTM as aggregation module for Static+avg+REG model}
    \label{table:lstm_sc}
\end{table*}

\begin{table*}
    \centering
    \begin{tabular}{|c|c|c|c|c|}
    \hline
    Model &  All 360 & Front 180 & Front Facing & Back\\
    \hline
    Squeezenet & 16.68 & 14.26 & 12.58 & 25.37 \\
    Shufflenet & 17.43 & 15.83 & 15.81 & 23.19 \\
    Mobilenet & 15.43 & 13.7 & 12.95 & 21.62 \\
    Resnet18 & 14.35 & 12.6 & 12.5 & 20.65 \\
    Hardnet68 & 13.75 & 12.14 & 12.07 & 19.54 \\
   \hline
    \end{tabular}
    \caption{Angular error with LSTM as aggregation module for Static+wavg+REG model}
    \label{table:lstm_wsc}
\end{table*}

\section{Performance Data on Different Backbones}
In Tables \ref{table:raw_all360},\ref{table:raw_front180}, \ref{table:raw_front40} and \ref{table:raw_back} we present the average angular error along with standard error of our different static models over multiple backbones. Note that for generating this data, each model configuration was trained three times independently and mean and the standard deviation of the three average angular error numbers are reported.

\clearpage
\bibliography{egbib}